\documentclass[letterpaper]{article} 
\usepackage{aaai2027}  
\usepackage[hyphens]{url}  
\usepackage{graphicx} 
\usepackage{natbib}  
\usepackage{caption} 
\usepackage{algorithm}
\usepackage{algorithmic}

\usepackage{newfloat}
\usepackage{amsmath}
\usepackage{amssymb}
\usepackage{multirow}
\usepackage{siunitx} 
\usepackage{listings}
\usepackage{enumitem} 
\DeclareCaptionStyle{ruled}{labelfont=normalfont,labelsep=colon,strut=off} 
\floatstyle{ruled}
\newfloat{listing}{tb}{lst}{}
\floatname{listing}{Listing}
\definecolor{NiceGreen}{RGB}{2,158,115} 
\definecolor{NiceRed}{RGB}{213,94,0}
\newcommand{\up}[1]{\textcolor{NiceGreen}{\scriptsize$\uparrow$#1}}
\newcommand{\down}[1]{\textcolor{NiceRed}{\scriptsize$\downarrow$#1}}
\usepackage{booktabs}

\newcommand{\Ev}[1]{\hyperref[ev:#1]{[A#1]}}
\usepackage{tikz}
\usetikzlibrary{shadows}
\definecolor{kleinblue}{RGB}{0, 47, 167} 
\definecolor{kleinblue2}{RGB}{20, 20, 125} 
\definecolor{green}{RGB}{146, 192, 113} 
\newcommand*\circled[1]{%
\tikz[baseline=(char.base)]{
  \node[shape=circle, fill=kleinblue2, draw=white, thick, drop shadow={shadow xshift=0.2ex,shadow yshift=-0.2ex}, inner sep=1pt] (char) {\textcolor{white}{#1}};
}}
\title{Grounding Isn't Knowing: Do VLMs Need Object Localization \\ for Spatial Reasoning?}
\author{
    Xiwei Liu\textsuperscript{\rm 1}, Yulong Li\textsuperscript{\rm 1}, Xinlin Zhuang\textsuperscript{\rm 1,2}, Xuhui Li\textsuperscript{\rm 1}, Zhixiang Lu\textsuperscript{\rm 3},\\ Haolin Yang\textsuperscript{\rm 1}, Imran Razzak\textsuperscript{\rm 1},
    Yutong Xie\textsuperscript{\rm 1}\corresponding
}
\affiliations{
    \textsuperscript{\rm 1}Mohamed bin Zayed University of Artificial Intelligence\\
    \textsuperscript{\rm 2}The Chinese University of Hong Kong, \textsuperscript{\rm 3}University Of Liverpool 

}

\begin{document}

\maketitle

\begin{abstract}
Vision-language models (VLMs) can answer spatial questions, yet the mechanisms connecting object grounding to spatial reasoning remain poorly understood. It is underexplored whether spatial reasoning internally requires precise objects localization, or can bypass explicit localization through global layout cues. In this work, we investigate two representative model families, LLaVA-1.5 and Qwen2.5-VL, using a suite of mechanistic interpretability tools, including token ablation, layer-wise probing, attention knockout, and causal mediation analysis. We find that spatial relation prediction follows a staged grounding-to-reasoning process in which object-aligned tokens establish coarse target-reference anchors, while precise bounding-box boundaries are not required. Positional information becomes decodable before relation decisions emerge, and a small set of attention heads mediates the causal effects of both localization and spatial reasoning. The two tasks share early grounding-related processing but ultimately rely on partially distinct specialized pathways. Through rigorous experiments, we provide a token-, layer-, and head-level account of how VLMs transform object grounding into spatial relations, showing that knowing where objects are is not equivalent to knowing how they relate.
\end{abstract}

\section{Introduction}
\label{sec:intro}

Spatial reasoning is essential for interacting with and reasoning about the visual world. Modern vision-language models (VLMs)~\cite{bai2025qwen3,comanici2025gemini} can answer questions about relative object positions, including whether one object is to the left of, behind, or in front of another~\cite{kamath2023s,chen2025spatial}. Yet a correct answer does not establish that the model has localized the queried objects or derived the relation from their positions. The same prediction may arise from object-specific evidence, coarse scene layout, contextual regularities, or language priors~\cite{li2025reading,schaumloffel2026mechanisms}. Benchmark accuracy therefore reveals what a model predicts, but provides limited insight into how object grounding is transformed into a spatial decision~\cite{tong2024eyes}.

Unlike text, an image is inherently organized across two spatial dimensions. Most VLMs nevertheless inherit a Transformer pipeline that partitions an image into patches and serializes them as a one-dimensional token sequence~\cite{dosovitskiy2020image}. Because sequence order does not explicitly preserve the image’s native two-dimensional topology, spatial structure must be recovered through positional encodings and interactions among visual tokens~\cite{chu2024visionllama,huang2025omega}. Fixed-resolution models commonly use learnable absolute position embeddings, whereas dynamic-resolution architectures can encode horizontal and vertical coordinates through two-dimensional rotary position embeddings (2D-RoPE)~\cite{huang2025positional}. These mechanisms preserve where visual tokens occur, but do not by themselves establish which tokens belong to the queried objects or how those objects relate~\cite{li2025reading,bousselham2024grounding,zhang2025cross,li2025token}.

Recent studies have strengthened explicit visual grounding in VLMs through localized visual tokenization, spatial prompting, grounded visual reasoning, and specialized localization heads~\cite{ma2024groma,dorkenwald2024pin,kang2025your}. Coordinate-based instruction tuning has also improved spatial-relation performance, suggesting a functional connection between localization supervision and spatial reasoning~\cite{ranasinghe2024learning}. However, these findings show that localization can be learned or exploited, not that a model must compute precise object locations before answering a relational query. Emerging evaluations further reveal that strong performance on conventional grounding benchmarks can coexist with severe failures on discriminative relational queries~\cite{li2026groundingme}. Whether localization is a prerequisite, a parallel capability, or a correlated outcome of spatial reasoning therefore remains underexplored.
\begin{figure*}[t]
    \centering
    \includegraphics[width=\linewidth]{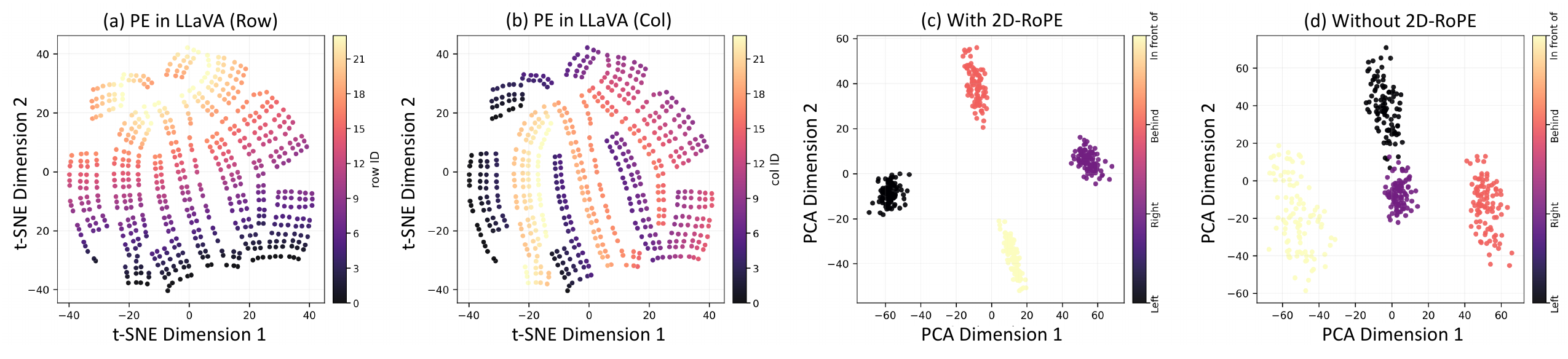}
    \vspace{-2em}
    \caption{(a) and (b): The geometry of the 1D absolute position embedding in LLaVA-1.5-7B. The visualization is performed via t-SNE for dimensionality reduction. The labels are set to row IDs and column IDs, respectively. (c) and (d): The PCA results of the direction vectors for left of / right of / in front of / behind with and without position embedding in Qwen2.5-7B.}
    \label{fig:position}
\end{figure*}

These advances expose a conceptual ambiguity: precise localization, relational grounding, and spatial reasoning are often evaluated as if they were interchangeable. Precise localization requires recovering object boundaries, relational grounding requires binding the target and reference to spatially anchored representations, and spatial reasoning requires transforming those representations into a directed relation. Recent training frameworks and benchmarks increasingly separate object localization from broader spatial understanding~\cite{li2025spatialladder,jia2025omnispatial}, while preference-based and token-based grounding methods optimize coordinate fidelity as a distinct objective~\cite{qiu2025spatial,ren2026grounding}. This distinction motivates our central question: must a VLM precisely localize both objects before predicting their relation, or can it reach the correct decision from coarser object-grounded layout signals?

To resolve this question, we investigate LLaVA-1.5 and Qwen2.5-VL across tokens, layers, and attention heads, combining object-region interventions, layer-wise probing, attention knockout, causal mediation, and targeted head ablation~\cite{neo2025towards,bi2025unveiling}. Together, these analyses reveal a consistent but nontrivial separation between precise localization and spatial reasoning in how object-grounded evidence is represented and routed. Our main findings are: 

\circled{1} \textbf{Precise localization is not required for spatial reasoning}. Removing object-interior evidence sharply degrades coordinate recovery while largely preserving directional judgments, showing that accurate bounding boxes are not a necessary intermediate for relation prediction. 

\circled{2} \textbf{Spatial reasoning relies more on coarse object layout than on exact boundary geometry}. Relation accuracy remains stable after removing fine boundary evidence but declines when interventions extend into the surrounding context, indicating that coarse object-centered layout provides a more relevant spatial signal than exact geometry. 

\circled{3} \textbf{Position precedes relation without forming a strict serial bottleneck.} Object positions become decodable in intermediate layers, whereas relation answers emerge later and may remain correct despite localization failures.

\circled{4} \textbf{Localization and spatial reasoning are mediated by sparsely overlapping attention circuits.} Object-grounded evidence is carried by a small set of shared heads, while being largely transformed through task-specific pathways.

\section{Background}\label{sec:background}

\paragraph{Sequential Visual Representation in VLMs.}
\label{sec:Sequential Visual Representation}
A vision-language model typically consists of a visual encoder $f_V$, a modality connector $f_C$, and a language model $f_L$. Given an image $\mathbf{X}_V$, the visual encoder first partitions it into an $H\times W$ grid of patches, $\mathbf{P}
=
(\mathbf{p}_1,\ldots,\mathbf{p}_{N_V})
\in\mathbb{R}^{N_V\times D_V}$,
where $N_V=HW$ and $D_V$ is the dimension of the visual encoder. The grid is then flattened in raster order, such that the patch at row $r$ and column $c$ is assigned the sequence index $i=rW+c$. The visual encoder and connector transform the patch sequence into language-compatible visual embeddings,
\begin{equation}
\mathbf{V}
=
f_C\!\left(f_V(\mathbf{P})\right)
\in\mathbb{R}^{N_V\times D_L}.
\end{equation}
Given text embeddings
$\mathbf{T}\in\mathbb{R}^{N_T\times D_L}$,
$f_L$ receives
\begin{equation}
\mathbf{H}^{(0)}
=
[\mathbf{V};\mathbf{T}]
\in
\mathbb{R}^{(N_V+N_T)\times D_L}.
\end{equation}

Although rasterization allows images to be processed as token sequences, it does not explicitly preserve their original two-dimensional organization. Positional encoding is therefore required to make attention computations sensitive to spatial structure
\cite{dosovitskiy2020image,liu2023visual}.

\paragraph{Position Encoding.}
\label{sec:position-encoding}
Position encoding in visual transformers can be broadly divided into absolute and relative. Given an input sequence
$\mathbf{X}=(\mathbf{x}_1,\ldots,\mathbf{x}_N)\in\mathbb{R}^{N\times D}$,
absolute position encoding assigns a vector $\mathbf{e}_i$ to each sequence position before being passed into the Transformer blocks:
\begin{equation}
\tilde{\mathbf{X}}
=
\mathbf{X}+\mathbf{E}
=
(\mathbf{x}_1+\mathbf{e}_1,\ldots,
 \mathbf{x}_N+\mathbf{e}_N).
\label{eq:absolute-pe}
\end{equation}
The position vectors may be fixed or learnable. For a fixed-resolution image, each sequence index corresponds to a persistent patch coordinate. Consequently, a one-dimensional position-embedding table can still encode the row and column structure of the original image grid.
Relative position encoding instead introduces the displacement between tokens into the attention computation. Rotary position embedding applies a position-dependent rotation $f(\cdot)$ to queries and keys
\cite{su2024roformer}. For query $\mathbf{q}_m$ and key $\mathbf{k}_n$ at positions $m$ and $n$, their inner product can be written as
\begin{equation}
\left\langle
f(\mathbf{q}_m,m),
f(\mathbf{k}_n,n)
\right\rangle
=
\operatorname{Re}
\left[
\mathbf{q}_m\mathbf{k}_n^{*}
e^{\mathrm{i}(m-n)\theta}
\right],
\label{eq:1d-rope}
\end{equation}
where $(\cdot)^{*}$ denotes complex conjugation and $\theta$ is a preset angular frequency. The resulting attention score depends explicitly on the relative displacement $m-n$.

For two-dimensional visual inputs, each token is assigned coordinates
$\mathbf{s}_i=(x_i,y_i)$.
The query and key representations are divided into horizontal and vertical subspaces, giving
\begin{equation}
\begin{aligned}
&
\left\langle
f(\mathbf{q}_i,x_i,y_i),
f(\mathbf{k}_j,x_j,y_j)
\right\rangle \\
&=
\operatorname{Re}
\left[
\mathbf{q}_i^{x}(\mathbf{k}_j^{x})^{*}
e^{\mathrm{i}(x_i-x_j)\theta}
+
\mathbf{q}_i^{y}(\mathbf{k}_j^{y})^{*}
e^{\mathrm{i}(y_i-y_j)\theta}
\right].
\end{aligned}
\label{eq:2d-rope}
\end{equation}
Thus, 2D RoPE represents spatial interactions through the relative displacement $\Delta\mathbf{s}_{ij}=(x_i-x_j,\;y_i-y_j)$,
allowing the same positional mechanism to ggeneralize consistently across varying image resolutions and sequence lengths.

In this work, we study LLaVA-1.5 and Qwen2.5-VL as representative instantiations of these two widely used mechanisms. LLaVA-1.5 uses learnable one-dimensional absolute position embeddings in its visual encoder
\cite{liu2023visual},
whereas Qwen2.5-VL employs 2D RoPE to support dynamic-resolution inputs
\cite{bai2025qwen25vltechnicalreport}.

\paragraph{From Positional Geometry to Object Grounding.}
\label{sec:position-to-grounding}
To verify how absolute and relative position encodings preserve spatial structure, we first apply t-SNE to reduce the dimensionality of the position embedding from LLaVA-1.5-7B to two dimensions. As visualized in Figure~\ref{fig:position}(a)(b), the geometric structure of the absolute position embedding exhibits distinct rows and columns. In the 2D-RoPE setting, we summarize an object $o$ by mean-pooling its aligned visual tokens,
\begin{equation}
\bar{\mathbf{h}}_o
=
\frac{1}{|\mathcal{S}_o|}
\sum_{i\in\mathcal{S}_o}
\mathbf{h}_i,
\end{equation}
and characterize a target-reference pair using the direction vector
$\mathbf{d}_{t,r}=\bar{\mathbf{h}}_t-\bar{\mathbf{h}}_r$.
Figure~\ref{fig:position}(c)(d) demonstrates the collinearity and orthogonality of the direction vectors, while the geometry breaks down without 2D RoPE.

These observations show that VLM representations preserve recoverable spatial geometry, but encoding spatial structure does not explain how it supports relation reasoning. A model may first recover object boundaries and compare them, or infer the relation directly from coarse object-centered cues. If localization is necessary, relation accuracy should decline when bounding-box localization fails. Otherwise, relation judgments may remain correct as long as coarse spatial grounding is preserved. We test this distinction through controlled interventions and trace the information across visual tokens, decoder layers, and attention heads.

\section{Preliminaries}
\subsection{Dataset and Task}\label{subsec:dataset}

To ensure that our analyses isolate the object-level visual evidence used by VLMs for spatial perception, we construct a carefully curated dataset derived from the What’s Up benchmark~\cite{kamath2023s}.

\paragraph{Base Dataset and Filtering.}
We use controlled real-world images from What’s Up Subsets A and B. Subset A contains objects positioned \textit{on}, \textit{under}, \textit{left of}, or \textit{right of} a table, chair, or armchair. We remove samples labeled \textit{on} or \textit{under} and retain only the left-right relations. Subset B contains pairs of household objects arranged \textit{in front of}, \textit{behind}, \textit{left of}, or \textit{right of} one another, all of which are retained. We further annotate the target and reference objects with bounding boxes and segmentation masks. After filtering, the dataset contains 1,228 object annotations across \num{614} images, including \num{206} images from Subset A and \num{408} images from Subset B.
\begin{figure}[t]
    \centering
    \includegraphics[width=\linewidth]{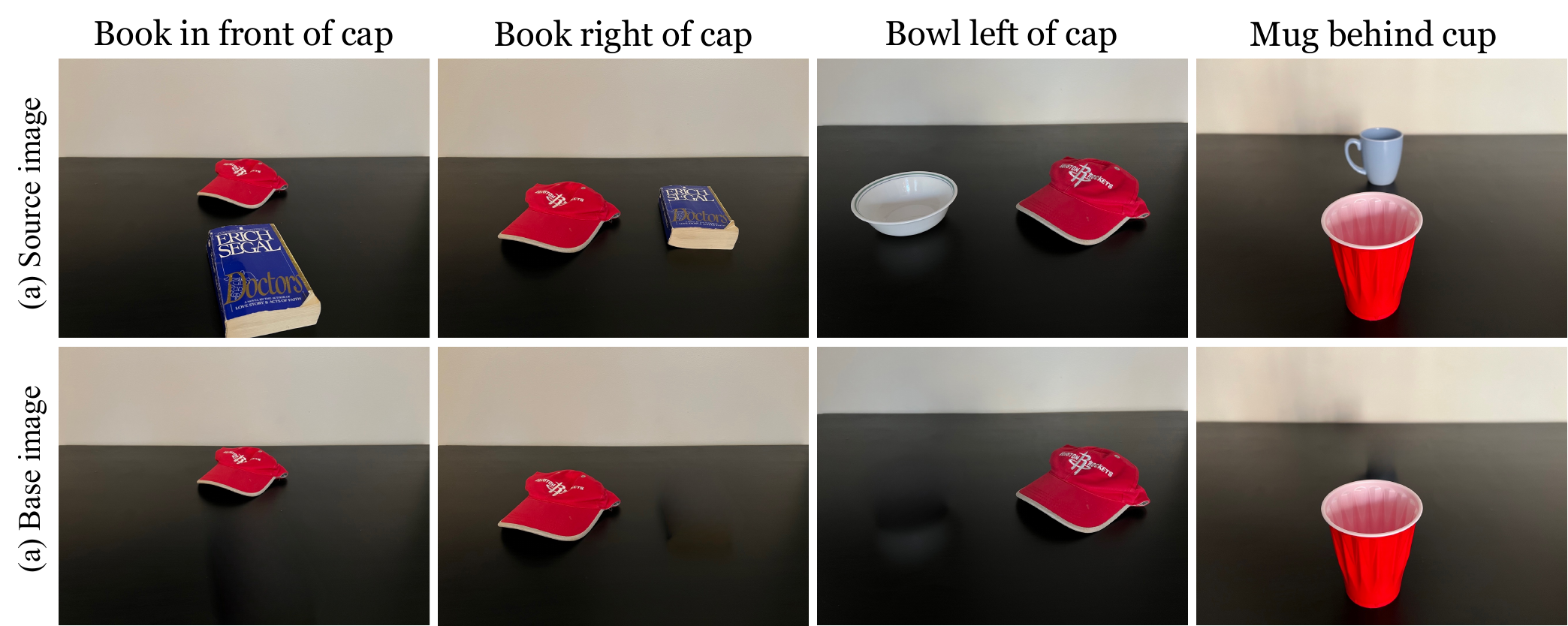}
    \caption{Examples in the curated What’s Up dataset.}
    \label{fig:data}
\end{figure}
\paragraph{Object-Removed Control Set.}
Contextual cues can lead to \textit{hallucinated detections}, where models predict the presence of an object solely from background context. To control for this effect, we construct an auxiliary object-removed variant of the dataset. For each image, the target object is removed and the missing region is inpainted using LaMa~\cite{suvorov2022resolution}, which reconstructs background structure with high realism. 
We retain only those image pairs for which a model correctly identifies the object in the original image but fails to do so in the inpainted counterpart. This ensures that subsequent analyses rely on real object evidence rather than contextual correlations. Examples of the inpainted dataset are provided in Figure~\ref{fig:data}. More details are shown in Supplement.\\
\begin{table*}[t]
\centering
\resizebox{\textwidth}{!}{%
\begin{tabular}{lcccc|ccc||cccc}
\toprule
Models: & \multicolumn{4}{c}{\textbf{LLaVA-1.5 7B}} & \multicolumn{3}{c}{\textbf{LLaVA-1.5 13B}} &\multicolumn{4}{c}{\textbf{Qwen2.5-VL 7B}} \\
\cmidrule(lr){1-1} \cmidrule(lr){2-5} \cmidrule(lr){6-8} \cmidrule(lr){9-12}
\textbf{\shortstack{Strategy}} 
& \multicolumn{1}{c}{\shortstack{Token (\%)}} 
& Loc-T (\%) 
& Loc-R (\%) 
& \multicolumn{1}{c|}{Rel. (\%)} 
& Loc-T (\%) 
& Loc-R (\%) 
& \multicolumn{1}{c|}{Rel. (\%)} 
& \multicolumn{1}{c}{\shortstack{Token (\%)}} 
& Loc-T (\%) 
& Loc-R (\%) 
& \multicolumn{1}{c}{Rel. (\%)} \\
\midrule

Baseline 
& \textit{0} & 10.15 & 14.81 & 24.82 
& 29.91 & 39.03 & 40.23 
& \textit{0} & 84.53 & 86.64 & 96.11 \\
\midrule

\textminus{} 1 Padding 
& \textit{1.77} & 8.79 \down{1.36} & 14.33 \down{0.48} & 24.61 \down{0.21} 
& 23.51 \down{6.40} & 39.25 \up{0.22} & 40.07 \down{0.16}
& \textit{2.24} & 84.58 \up{0.05} & 86.75 \up{0.11} & 95.93 \down{0.18} \\

\textbf{Target} 
& \textit{4.94} & 4.78 \down{5.37} & 14.71 \down{0.10} & 24.59 \down{0.23} 
& 2.01 \down{27.90} & 39.90 \up{0.87} & 40.05 \down{0.18}
& \textit{4.23} & 19.60 \down{64.93} & 86.32 \down{0.32} & 94.79 \down{1.32} \\

+ 1 Padding 
& \textit{9.52} & 0.38 \down{9.77} & 17.59 \up{2.78} & 24.57 \down{0.25} 
& 0.12 \down{29.79} & 36.91 \down{2.12} & 27.85 \down{12.38}
& \textit{6.71} & 2.55 \down{81.98} & 84.69 \down{1.95} & 89.74 \down{6.37} \\
\midrule

\textminus{} 1 Padding 
& \textit{2.59} & 10.53 \up{0.38} & 14.12 \down{0.69} & 24.75 \down{0.07} 
& 29.64 \down{0.27} & 31.38 \down{7.65} & 39.90 \down{0.33}
& \textit{3.26} & 84.26 \down{0.27} & 88.60 \up{1.96} & 96.42 \up{0.31} \\

\textbf{Reference} 
& \textit{7.64} & 10.58 \up{0.43} & 7.11 \down{7.70} & 24.59 \down{0.23} 
& 29.15 \down{0.76} & 2.33 \down{36.70} & 39.41 \down{0.82}
& \textit{6.46} & 84.31 \down{0.22} & 25.57 \down{61.07} & 95.77 \down{0.34} \\

+ 1 Padding 
& \textit{14.33} & 10.91 \up{0.76} & 0.11 \down{14.70} & 24.42 \down{0.40} 
& 27.69 \down{2.22} & 0.24 \down{38.79} & 36.32 \down{3.91}
& \textit{10.16} & 83.28 \down{1.25} & 4.34 \down{82.30} & 92.83 \down{3.28} \\
\midrule

\multirow{6}{*}{\shortstack{Random\\(3 seeds)}} 
& \textit{1.77} & 10.23 \up{0.08} & 14.72 \down{0.09} & 24.88 \up{0.06} 
& 29.74 \down{0.17} & 39.43 \up{0.40} & 40.28 \up{0.05}
& \textit{2.24} & 84.46 \down{0.07} & 86.51 \down{0.13} & 95.97 \down{0.14} \\

& \textit{4.94} & 9.75 \down{0.40} & 16.85 \up{2.04} & 24.63 \down{0.19} 
& 29.38 \down{0.53} & 39.18 \up{0.15} & 39.92 \down{0.31}
& \textit{4.23} & 84.35 \down{0.18} & 86.47 \down{0.17} & 95.38 \down{0.73} \\

& \textit{9.52} & 9.38 \down{0.77} & 13.26 \down{1.55} & 24.51 \down{0.31} 
& 29.15 \down{0.76} & 38.96 \down{0.07} & 39.27 \down{0.96}
& \textit{6.71} & 83.29 \down{1.24} & 85.71 \down{0.93} & 94.51 \down{1.60} \\

& \textit{2.59} & 10.12 \down{0.03} & 15.11 \up{0.30} & 24.68 \down{0.14} 
& 29.69 \down{0.22} & 39.52 \up{0.49} & 39.96 \down{0.27}
& \textit{3.26} & 84.22 \down{0.31} & 86.34 \down{0.30} & 95.85 \down{0.26} \\

& \textit{7.64} & 10.37 \up{0.22} & 14.38 \down{0.43} & 24.77 \down{0.05} 
& 29.26 \down{0.65} & 38.94 \down{0.09} & 39.54 \down{0.69}
& \textit{6.46} & 83.96 \down{0.57} & 84.68 \down{1.96} & 95.03 \down{1.08} \\

& \textit{14.33} & 9.51 \down{0.64} & 14.89 \up{0.08} & 24.47 \down{0.35} 
& 28.88 \down{1.03} & 38.71 \down{0.32} & 39.06 \down{1.17}
& \textit{10.16} & 83.64 \down{0.89} & 83.88 \down{2.76} & 95.12 \down{0.99} \\
\bottomrule
\end{tabular}
}
\caption{Performance under token ablation. Baseline removes no token. Target and reference ablations replace visual tokens selected from the corresponding object masks with a global average visual embedding; padding variants shrink or dilate the object-token mask by one token. Token (\%) is the average percentage of image tokens replaced. Loc-T (Target) and Loc-R (Reference) report localization accuracy averaged over IoU$@$0.5/0.7/0.9, and Rel. reports spatial relation accuracy.}
\label{tab:ablation}
\end{table*}
We evaluate models on two complementary visual tasks: localizing the target and reference objects and predicting their spatial relation. For each task, we design a different prompt for the same image. See detailed prompts in Supplement.

\paragraph{Localization.}
We prompt the model to predict target and reference bounding boxes separately. Each parsed prediction is compared with its ground-truth box using intersection over union (IoU). We report target and reference localization as the mean success rate at IoU thresholds of 0.5, 0.7, and 0.9.

\paragraph{Spatial Reasoning.}
We use the original four-way multiple-choice protocol from What’s Up, preserving its subset-specific distractors. Accuracy is the proportion of predicted target-to-reference relations that match the ground-truth.

\section{Experiments}
We examine localization and spatial reasoning at progressively finer scales. Token interventions test whether object evidence is necessary for relation prediction, layer-wise analyses trace when relevant information emerges and is used, and head-level mediation and ablation identify the attention components that route grounded evidence into outputs.
\subsection{Visual Information Ablation}
We conduct an ablation study to investigate the contribution of visual input tokens to the performance of the VLMs.  

\paragraph{Method.} We ablate visual information at the LLM input, i.e., after the multimodal projection but before positional encodings and autoregressive processing. To remove image-specific content while preserving domain-consistent embedding statistics, we replace the original visual token embeddings with a global average visual embedding computed once over the ImageNet~\cite{deng2009imagenet} validation set.

\begin{figure*}[t]
    \centering
    \includegraphics[width=\linewidth]{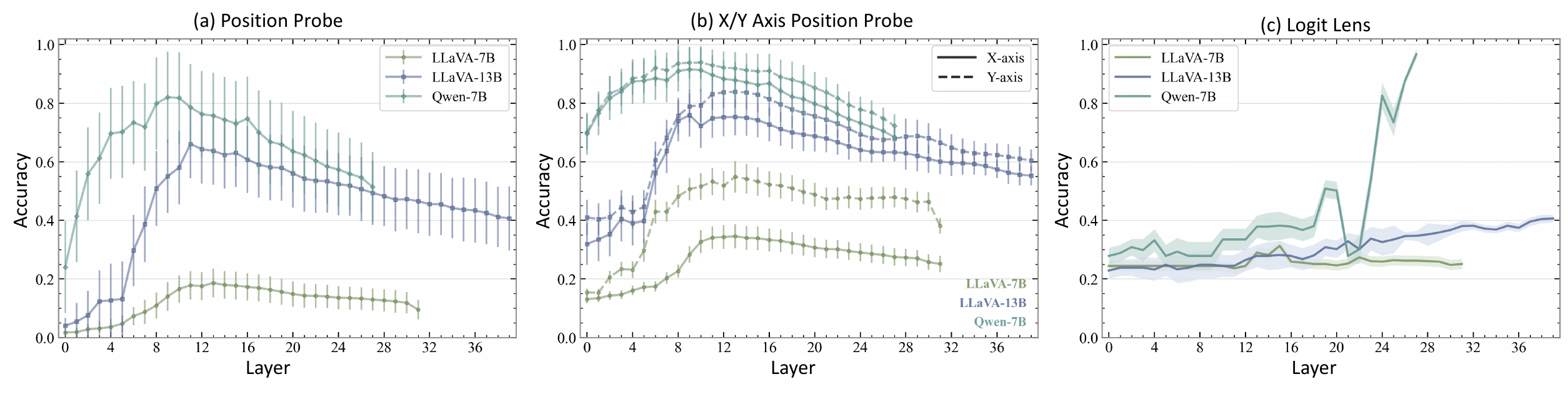}
    \vspace{-2em}
    \caption{Layer-wise positional decoding and logit lens. (a) Joint two-dimensional position-probe accuracy across the LLM decoder layers of LLaVA-1.5-7B, LLaVA-1.5-13B, and Qwen2.5-VL-7B. (b) Axis-specific position-probe accuracy. (c) Spatial-relation answer accuracy obtained at each layer by applying a training-free logit lens to the final prompt-token representation. Error bars and shaded regions denote the standard error across samples.}
    \label{fig:position probe}
    \vspace{-1em}
\end{figure*}

\begin{figure*}[ht]
    \centering
    \includegraphics[width=\linewidth]{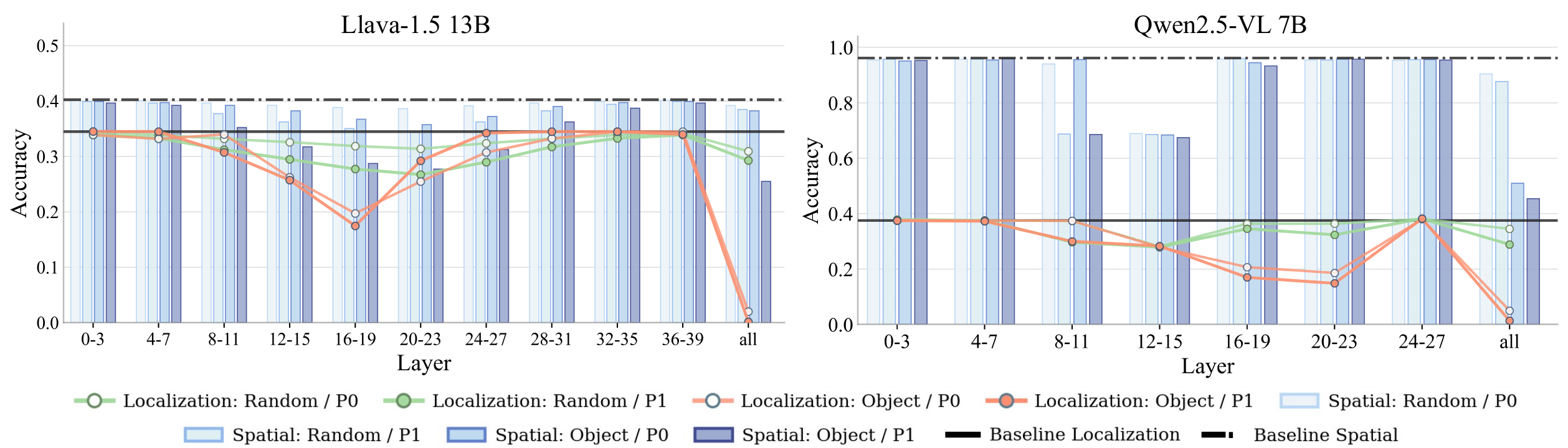}
    \caption{Performance after attention knockout. We block attention to object-aligned visual tokens within grouped decoder layers and across all layers. Lines and bars report localization and spatial-relation accuracy, respectively, under the Padding=0 (P0) and Padding=1 (P1) settings. Object-token knockout is compared with a size-matched random-token control. Solid and dash-dotted horizontal lines indicate the localization and spatial-relation accuracy of the unmodified model.}
    \label{fig:knockout}
\end{figure*}

\noindent\textit{(i) Object Tokens}: We rasterize each segmentation mask onto the visual-token grid and select all tokens that overlap with it by at least one pixel. To probe for boundary sensitivity and context dependence, we shrink or dilate the mask by 1 or 2 token padding. This produces nested regions ranging from the object interior to its surrounding context.

\noindent\textit{(ii) Random Tokens}: As a control, we replace an equal number of randomly selected visual tokens for each object-aligned intervention. Results are averaged over three seeds.

\paragraph{Result.} 
Table~\ref{tab:ablation} shows that object-aligned tokens selectively support localization. Target-token ablation reduces Loc-T by 5.37, 27.90, and 64.93 points for LLaVA-7B, LLaVA-13B, and Qwen-7B, respectively, while reference-token ablation reduces Loc-R by 7.70, 36.70, and 61.07 points. Localization of the unablated object remains stable, and matched random removal causes substantially smaller changes. Yet relation accuracy drops by at most 1.32 points under the original masks despite large localization losses. Mask dilation increasingly impairs relation prediction, whereas erosion produces weaker localization effects. Thus, precise localization depends on object-internal evidence, while relation prediction remains robust until the intervention reaches the surrounding context.

\subsection{Position Decoding}
Token ablation shows relation prediction can survive the loss of evidence required for localization. We next ask whether this dissociation is reflected in when positional and relation information becomes identifiable across decoder depth.

\paragraph{Method.}
To evaluate positional identifiability, we train separate linear classifiers at each decoder layer to predict the two-dimensional grid position of every image token. The horizontal (x) and vertical (y) coordinates are predicted independently, and a joint prediction is correct only when both coordinates are recovered. The classifiers are trained for 10 epochs on 50,000 ImageNet images and evaluated on the What’s Up images. To trace when relation-specific information becomes accessible to the model’s output, we further employ the logit lens, a training-free method for interpreting intermediate activations~\cite{logitlens}. At each layer, the final prompt-token representation is passed through the model’s final normalization and language-model head. We then compare the logits assigned to the answer options.

\paragraph{Results.}
Figure~\ref{fig:position probe} reveals that positional information increases rapidly, peaking around layer 14 for LLaVA-7B, layer 12 for LLaVA-13B, and layer 10 for Qwen-7B before gradually declining. The axis-specific probes show consistent trends, confirming that both horizontal and vertical coordinates remain recoverable across substantial portions of the decoder. In contrast, the logit lens reveals markedly different answer-formation dynamics. Qwen2.5-VL remains near chance in early layers but rises sharply after layer 19, reaching approximately 97\% accuracy at the final layer. LLaVA-7B remains near chance throughout, whereas LLaVA-13B improves gradually to approximately 40\%. These results show that positional decodability alone does not guarantee correct spatial reasoning. The temporal separation between position recovery and answer emergence is consistent with a staged process in which spatial layouts are represented before being transformed into relation-specific decisions.

\begin{figure*}[ht]
    \centering
    \includegraphics[width=\linewidth]{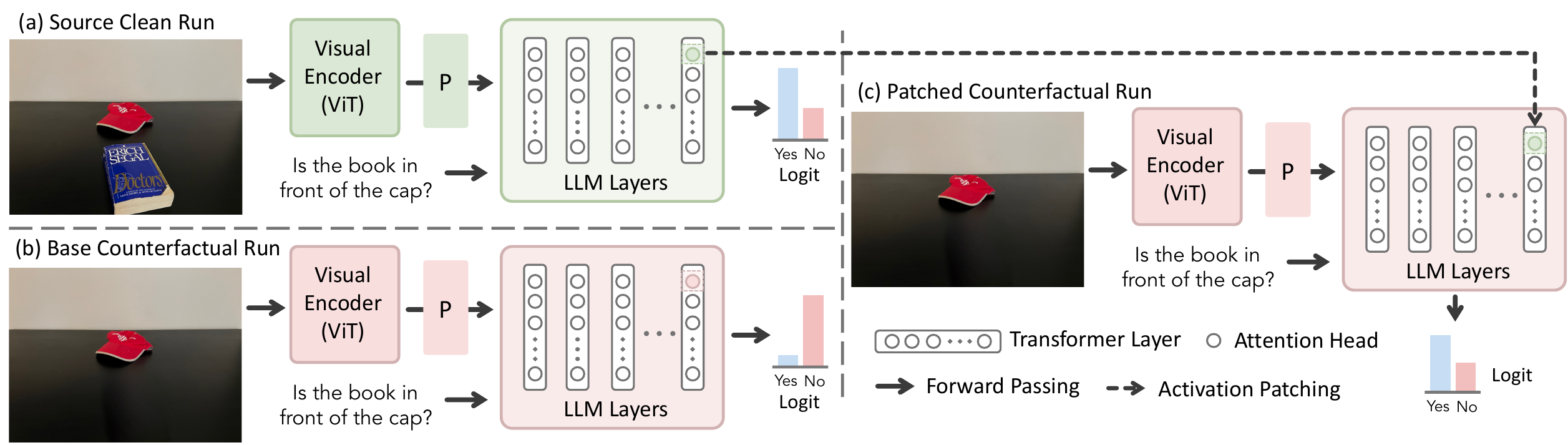}
    \vspace{-1em}
    \caption{Causal mediation via activation patching. We compare three model runs: (a) the source clean run, where the object is present and the model produces the correct answer; (b) the base counterfactual run, where the target object is removed and the model fails; and (c) the patched counterfactual run, where we transfer hidden activations from a selected attention head in the source run into the base run. Improvements in the patched prediction indicate that the transferred head carries task-relevant information. This procedure is repeated head-wise to quantify each head’s causal contribution.}
    \label{fig:cma visualize}
\end{figure*}
\begin{figure*}[ht]
    \centering
    \includegraphics[width=\linewidth]{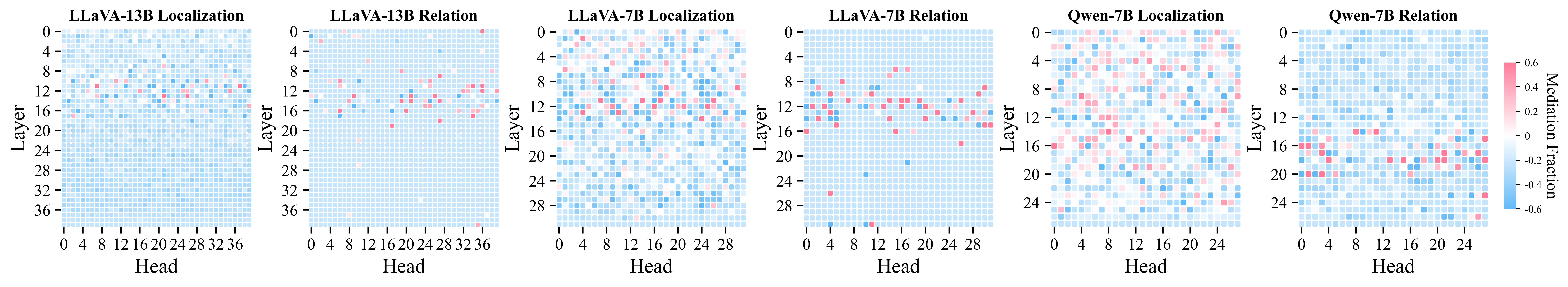}
    \vspace{-2em}
    \caption{Head-level causal mediation analysis. Mediation Fraction (MF) scores for every attention head across the decoder layers, shown separately for the localization and spatial reasoning tasks.}
    \label{fig:cma}
    \vspace{-1em}
\end{figure*}
\subsection{Attention Knockout}
Layer-wise probes reveal when spatial information is accessible, but not whether later computations actually use it. We therefore investigate where the model extracts and processes object-grounded visual information.
\paragraph{Method.}
We apply \textit{attention knockout}~\cite{geva-etal-2023-dissecting, neo2025towards} to block attention from all post-image tokens to object-aligned visual tokens. This intervention prevents object information from being extracted within the selected layers. We group consecutive layers into groups of four and block all attention heads within each group. We also perform knockout across all layers as a global intervention. We measure the resulting changes in localization and spatial-relation accuracy using our curated What's UP subset.

\paragraph{Results.}
The results are shown in Figure~\ref{fig:knockout}. For LLaVA-1.5-13B, object-token knockout produces the largest localization decline in layers 16--19, while spatial-relation accuracy remains comparatively stable under most local interventions. Blocking attention across all layers reduces localization to nearly zero and lowers relation accuracy toward chance level.
For Qwen2.5-VL-7B, localization is most sensitive to interventions in layers 16--23. Spatial-relation accuracy is also affected by several early and middle layer groups. Global object-token knockout again produces the strongest degradation in both tasks and is more disruptive than the matched random control. The similar trends under Padding = 0/1 indicate that these effects are not determined by a single choice of object-token boundaries.
These results reveal only partial overlap between the layers supporting localization and spatial-relation prediction. Precise localization depends most strongly on specific intermediate layers, whereas relation prediction draws on information distributed across a broader processing range. Nevertheless, the global intervention shows that spatial reasoning still depends on information extracted from object-aligned tokens. This distinction motivates our subsequent head-level analysis of the pathways that transform object grounding into spatial decisions.
\begin{figure*}[ht]
    \centering
    \includegraphics[width=\linewidth]{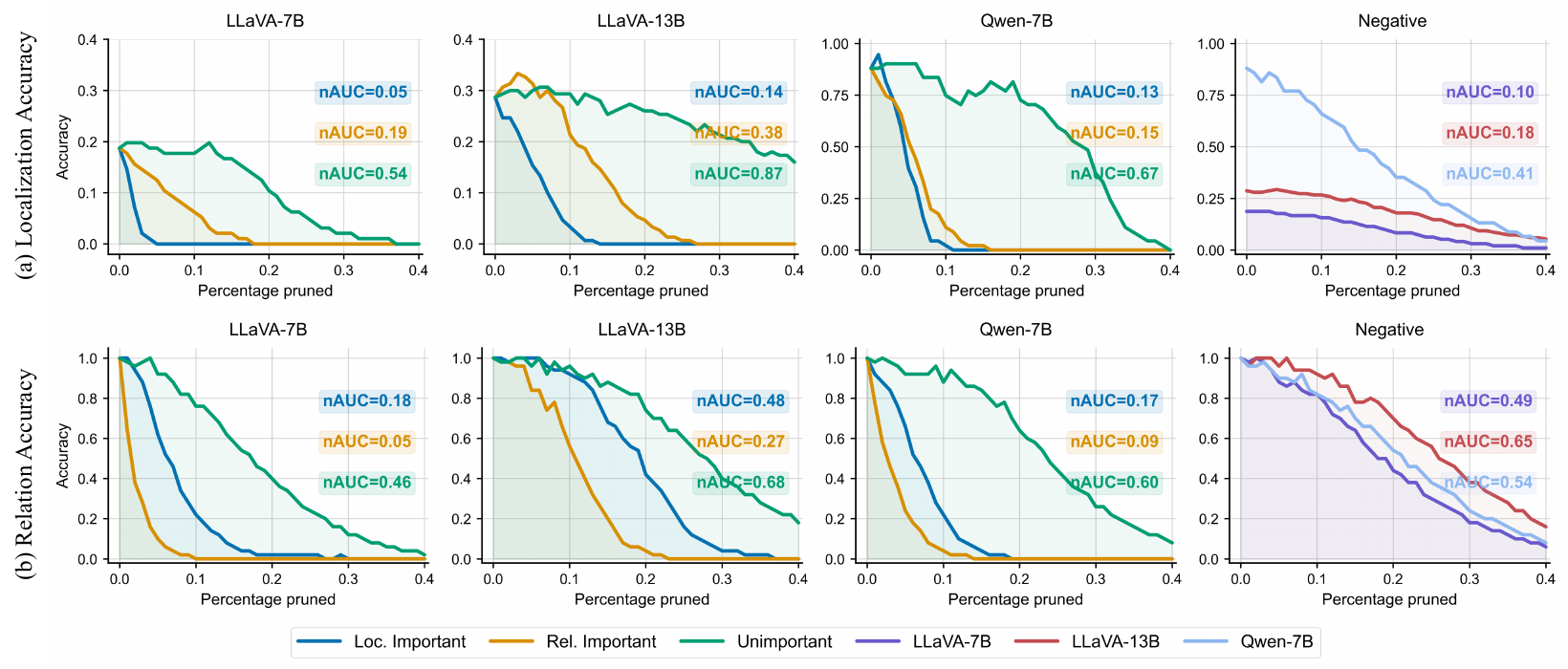}
    \vspace{-2em}
    \caption{Performance under cumulative ablation of attention heads. (a) Localization and (b) spatial-relation accuracy as increasing fractions of localization-important, relation-important, unimportant, and negative-MF heads are ablated. }
    \label{fig:head}
\end{figure*}
\subsection{Causal Mediation Analysis}
Attention-Knockout presents evidence that the mechanisms underlying localization and spatial reasoning are concentrated within a narrow region of the language model’s processing pipeline. We therefore apply causal mediation analysis (CMA)~\cite{yang2025emergent, wang2023interpretability, Meng2022factual} using activation patching to identify which attention heads causally contribute to solving the visual task.

\paragraph{Method.}
For each model, we curate up to 50 source-base pairs. The source image contains the target object, while the base image removes it using LaMa inpainting (Figure~\ref{fig:data}). For each attention head, we patch its source prompt-token activations into the corresponding base run. The counterfactual output measures whether the patched head restores information lost through object removal (Figure~\ref{fig:cma visualize}).

All runs are evaluated under teacher forcing using mean token-level negative log-likelihood (NLL). We score only numerical coordinate tokens for localization and the ground-truth answer token for a binary spatial-relation query. Let $L_{\mathrm{src}}$, $L_{\mathrm{base}}$, and $L_{\mathrm{patched}}$ denote the corresponding NLL values. We define the Mediation Fraction as
\begin{align}
    \mathrm{MF}
    =
    \frac{L_{\mathrm{base}}-L_{\mathrm{patched}}}
    {L_{\mathrm{base}}-L_{\mathrm{src}}}.
\end{align}
MF measures the fraction of the source-base gap restored by the patched head. MF $\approx$ 1 indicates that the patched head recovers most of the source-base performance gap; MF $\approx$ 0 indicates little contribution, while MF $<$ 0 indicates misleading or interfering effects. To limit affirmative bias in the binary relation task, we retain only pairs where the source prediction is correct and object removal degrades the answer. This paired design ensures that NLL changes reflect the recovery of object-dependent relational evidence.

\paragraph{Results.}
Figure~\ref{fig:cma} reports head-level MF scores. Consistent with attention knockout, high-MF heads concentrate in the early–mid layers of LLaVA-1.5 and the mid-late layers of Qwen2.5-VL. In LLaVA-1.5-13B, both tasks show their strongest effects around layers 11-16, but rely on substantially different heads within this shared range. In Qwen2.5-VL-7B, localization effects are weaker and more dispersed, whereas relation mediation concentrates around layers 16-20. Most heads outside these regions have MF scores near zero.

Mediation is sparse and largely task-specific. Among the top 10 heads, localization and relation prediction share only 3 heads in LLaVA-1.5-13B and 2 heads each in LLaVA-1.5-7B and Qwen2.5-VL-7B. The corresponding overlaps among the top 50 heads are only 3, 7, and 3. Thus, apparent overlap at the layer level masks substantial specialization at the head level. Negative MF values across all models further suggest that some isolated source activations are incompatible with the base-image computation and interfere with the correct prediction. Overall, localization and relation prediction reuse limited intermediate computation, but their dominant causal effects are mediated by largely distinct attention heads.

\subsection{Head Ablation Analysis}
\label{sec:head_ablation}
CMA reveals a small number of attention heads with high MF scores per task. To determine whether these heads are necessary rather than merely correlated with performance, we progressively remove them in a cumulative ablation study.

\paragraph{Method.}
We rank attention heads separately by their mean MF for localization and spatial relations. As controls, low-importance heads have the smallest mean absolute MF, while negative heads have the lowest signed MF across both tasks. We progressively
ablate a proportion of the total number of heads by setting
their outputs to zero during the forward pass and evaluate localization and binary relation accuracy. Each pruning curve is summarized by normalized AUC (nAUC), where lower values indicate greater sensitivity.

\paragraph{Results.}
Figure~\ref{fig:head} shows that removing task-important heads is substantially more damaging than removing low-importance heads. Localization declines fastest under the localization ranking, whereas relation accuracy is most sensitive to relation-important heads, confirming that CMA identifies functionally necessary components. Relation accuracy also drops strongly when localization-important heads are removed, indicating that both tasks reuse object-grounded intermediate representations despite limited head overlap. Ablating negative-MF heads produces gradual, model-dependent declines rather than consistent improvements. Negative mediation therefore reflects incompatibility during isolated activation patching, not necessarily a harmful role during normal inference. Together with attention knockout, these results support a staged mechanism in which shared object-grounded representations are transformed by largely specialized heads into precise locations or spatial decisions.

\section{Related Work}

\paragraph{Spatial Reasoning in VLMs.}
Spatial reasoning has been evaluated from complementary perspectives. VSR examines a broad range of relations in natural images, while BLINK isolates perception-intensive abilities such as relative depth, visual correspondence, and multi-view reasoning~\cite{fu2024blink}. TopViewRS further separates object recognition, localization, and relational inference in controlled top-view environments~\cite{li2024topviewrs}. Beyond evaluation, SpatialVLM introduces large-scale 3D supervision for qualitative and quantitative spatial reasoning~\cite{chen2024spatialvlm}, whereas SpatialRGPT incorporates region representations and depth cues to support grounded relational judgments~\cite{cheng2024spatialrgpt}. These studies establish that spatial reasoning remains challenging and can benefit from explicit spatial supervision. However, neither benchmark performance nor training gains determine whether a model must recover precise object locations before predicting a relation. 

\paragraph{Explicit Grounding in Multimodal Models.}
A parallel line of work equips multimodal models with explicit object grounding. Kosmos-2~\cite{peng2023kosmos2groundingmultimodallarge} and Shikra~\cite{chen2023shikra} generate coordinates directly within the language sequence, while Ferret combines discrete coordinates with continuous region features to support multi-granularity referring and grounding~\cite{you2024ferret}. GPT4RoI~\cite{zhang2025gpt4roiinstructiontuninglarge} and PVIT~\cite{chen2023position} inject region-level features into visual instructions. More recent systems extend grounding from bounding boxes to pixel-level masks through reasoning segmentation, grounded conversation, and multi-object segmentation~\cite{lai2024lisa,rasheed2024glamm,ren2024pixellm,zhang2024groundhog,li2025reading,schaumloffel2026mechanisms}. Collectively, these methods show that precise localization can be learned, elicited, or decoded from VLM representations, but not that such precision is causally required for spatial reasoning, since grounding supervision and output interfaces are introduced jointly.

\paragraph{Mechanistic Analysis of Multimodal Computation.}
Recent interpretability studies have begun to trace how visual evidence is transformed inside VLMs. Attention-lens analysis identifies distinct stages of visual enrichment and semantic refinement in intermediate layers~\cite{jiang2025devils}, while layer-wise information decomposition characterizes the transition from visual evidence to language-dominated predictions~\cite{li2025reading,wu2026vision}. Sparse feature analysis connects latent visual concepts to their downstream use in generated answers~\cite{lim2026visualscratchpad}, and head-level interventions reveal that cross-modal retrieval can be concentrated in a small fraction of attention heads~\cite{sun2026mechanistic}. These findings suggest that visual computation is both depth-dependent and functionally sparse. Most existing analyses, however, only explain a single output behavior such as recognition, retrieval, or hallucination.

\section{Conclusion}
Do VLMs really need object localization for spatial reasoning? Our results provide a qualified answer: VLMs require object grounding, but not precise localization. Spatial relations can survive severe localization failures, provided that coarse object-centered layout remains available. Thus, grounding is necessary evidence, but knowing exact object boundaries is not equivalent to knowing how objects relate. Whether
this mechanism generalizes to crowded scenes, 3D relations, and
temporal reasoning remains open.

\clearpage
\newpage
\bibliography{aaai2027}

\end{document}